\pdfoutput=1
\documentclass{article}
\usepackage{times}
\usepackage{url}
\usepackage{multirow}
\usepackage{graphicx}
\usepackage{subfig}
\usepackage{xcolor}

\title{Predictive Linguistic Features of Schizophrenia}

\author
       {\normalsize Efsun Sarioglu Kayi$^1$, Mona Diab$^2$, Luca Pauselli$^3$, Michael Compton$^4$, Glen Coppersmith$^3$
       \\
       $^1${\normalsize Department of Computer Science, Columbia University}\\
       $^2${\normalsize Department of Computer Science, George Washington University}\\
       $^3${\normalsize Medical Center, Columbia University}\\
       $^4${\normalsize Qntfy}\\
       }

\begin{document}

\maketitle

\begin{abstract}
Schizophrenia is one of the most disabling and difficult to treat of all human medical/health conditions, ranking in the top ten causes of disability worldwide. 
It has been a puzzle in part due to difficulty in identifying its basic, fundamental components.  
Several studies have shown that some manifestations of schizophrenia (e.g., the \emph{negative symptoms} that include blunting of speech prosody, as well as the \emph{disorganization} symptoms that lead to disordered language) can be understood from the perspective of linguistics. 
However, schizophrenia research has not kept pace with technologies in computational linguistics, especially in semantics and pragmatics. As such, we examine the writings of schizophrenia patients analyzing their syntax, semantics and pragmatics. In addition, we analyze tweets of (self proclaimed) schizophrenia patients who publicly discuss their diagnoses. For writing samples dataset, syntactic features are found to be the most successful in classification whereas for the less structured Twitter dataset, a combination of  features performed the best. 
\end{abstract}

\section{Introduction}
Schizophrenia is an etiologically complex, heterogeneous, and chronic disorder. It imposes major impairments on affected individuals, can be devastating to families, and it diminishes the productivity of communities. Furthermore, schizophrenia is associated with remarkably high direct and indirect health care costs.
Persons with schizophrenia often have multiple medical comorbidities, 
have a tragically reduced life expectancy, 
and are often treated without the benefits of sophisticated \emph{measurement-based care}.

Similar to other psychoses, schizophrenia has been studied extensively on the neurological and behavioral levels. Covingtion et al. \cite{Covington:2005aa} note the existence of many language abnormalities (in syntactic, semantic, pragmatic, and phonetic domains of linguistics) comparing patients to controls. They observed the following: 
\begin{itemize}
\item reduction in syntax complexity \cite{Fraser:1986aa}; 
\item impaired semantics, such as the organization of individual propositions into larger structures \cite{Rodriguez-Ferrera:2001aa}; 
\item abnormalities in pragmatics 
which is a level obviously disordered in schizophrenia
\cite{Covington:2005aa}; 
\item phonetic anomalies like flattened intonation (aprosody), more pauses, and constricted pitch/timbre \cite{Stein:1993aa}.
\end{itemize}
A few studies have used computational methods to assess acoustic parameters (e.g., pauses, prosody) that correlate with negative symptoms, but schizophrenia research has not kept pace with technologies in computational linguistics, especially in semantics and pragmatics. Accordingly, we analyze the predictive power of linguistic features in a comprehensive manner by computing and analyzing many syntactic, semantic and pragmatic features.
This sort of analysis is particularly useful for finding meaningful signals that help us better understand the mental health conditions.
To this end, we compute part-of-speech (POS) tags and dependency parses to capture the syntactic information in patients' writings. For semantics, we derive topic based representations and semantic role labels of writings. In addition, we add more semantics by adding dense features using clusters that are trained on online resources. For pragmatics, we consider the sentiment that exists in writings, i.e. \emph{positive} vs. \emph{negative} and its intensity. Finally, we explore the committed belief modality in an attempt to capture the commitment level of the writer to the propositions in their writing. 
To the best of our knowledge, no previous work has conducted comprehensive analysis of schizophrenia patients' writings from the perspective of syntax, semantics and pragmatics, collectively.
\section{Predictive Linguistic Features of Schizophrenia}
\label{sect:prediction}
\subsection{Dataset}
\label{ssect:dataset}
The first dataset called LabWriting consists of 93 patients with schizophrenia who were recruited from sites in both Washington, D.C. and New York City. This includes patients that have a diagnosis of schizophreniform disorder or first-episode or early-course patients with a psychotic disorder not otherwise specified. All patients were native English-speaking patients, aged 18-50 years and cognitively intact enough to understand and participate in the study. 

A total of 95 eligible controls were also native English speakers aged 18-50. Patients and controls did not differ by age, race, or marital status, however, patients were more likely to be male and had completed fewer years of education. All study participants were assessed for their ability to give consent, and written informed consent was obtained using Institutional Review Board-approved processes. 

Patients and controls were asked to write two paragraph-length essays: one about their average Sunday and the second about what makes them the angriest. The total number of writing samples collected from both patients and controls is 373. Below is a sample response from this dataset (text from patients rendered verbatim as is including typos):
\\
\\
\emph{The one thing that probably makes me the most angry is when good people receive the bad end of the draw. This includes a child being struck for no good reason. A person who is killed but was an innocent bystander. Or even when people pour their heart and soul into a job which pays them peanurs but they cannot sustain themselves without this income. Just in generul a good person getting the raw end of deal. For instance people getting laid off because their company made bad investments. the Higher ups keep their jobs while the worker ants get disposed of. How about people who take advantage of others and build an Empire off it like insurance or drug companies. All these good decent people not getting what they deserved. Yup that makes me angry.}
\\

In addition, we evaluated social media messages with  self-reported diagnoses of schizophrenia using the Twitter API. This dataset includes 174 users with apparently genuine self-stated diagnosis of a schizophrenia-related condition and 174 age and gender matched controls. Schizophrenia users were selected via regular expression on \emph{schizo} for a close phonetic approximation. Each diagnosis was examined by a human annotator to verify that it seems genuine. For each schizophrenia user, a control that had the same gender label and was closest in age was selected. The average number of tweets per user is around 2,800. Detailed information on this dataset can be found in \cite{mitchell-hollingshead-coppersmith:2015:CLPsych}. Below are some tweets from this dataset (they have been rephrased to preserve anonymity):
\\
\\
\emph{this is my first time being unemployed. please forgive me. i'm crazy. \#schizophrenia}
\\
\\
\emph{ i'm in my late 50s. i worry if i have much time left as they say people with \#schizophrenia die 15-20 years younger}
\\
\\
\emph{\#schizophrenia takes me to devil-like places in my mind }
\subsection{Approach and Experimental Design}
\label{sect:approach}
We cast the problem as a supervised binary classification task where a system should discriminate between a patient and a control. To classify schizophrenia patients from controls, we trained  support vector machines (SVM) with linear kernel and Random Forest classifiers. We used Weka \cite{Hall:2009:WDM:1656274.1656278} to conduct the experiments with 10-fold stratified cross validation. We report  F-Score, and Area Under Curve (AUC) value which is the area under receiver operating characteristics curve (ROC). 
\subsubsection{Syntactic Features}
To capture the syntactic information from writings, we produce the POS tags and dependency parse trees using Stanford Core NLP \cite{manning-EtAl:2014:P14-5}. To use these as features to the classifier, we calculate the frequency of each POS tag and dependencies from parse trees. 
For the Twitter dataset, we use a parser \cite{kong2014dependency} and POS tagger \cite{Gimpel:2011:PTT:2002736.2002747} that are specifically trained for social media data. 
\subsubsection{Semantic Features}
To analyze the semantics of the writings, we consider several sources of information. As a first approach, we use semantic role labeling (SRL). Specifically, we use Semafor \cite{Das:2010:PFP:1857999.1858136} tool to generate semantic role labels of the writings and then calculate the frequency of the labels as features for the classifier. For Twitter dataset, due to its short form and poor syntax, we were not able to compute SRL features.

In addition to SRL, we analyzed the topic distribution of writings using Latent Dirichlet Allocation (LDA) \cite{Blei:2003:LDA:944919.944937}. With this approach, we want to see the possibility of different themes emerging in the writings of patients vs. controls.  Using LDA, we represent each writing as a topic distribution where each topic is automatically learned as a distribution over the words of the vocabulary. We use the MALLET tool \cite{McCallumMALLET} to train the topic model and empirically choose number of topics based on best classification performance on a validation set. The best performing number of topics is 20 for LabWriting dataset and 40 for Twitter dataset.

Finally, we compute dense semantic features by computing clusters based on global word vectors. Specifically, for LabWriting dataset, we use word vectors trained on Wikipedia 2014 dump and Gigaword 5 \cite{gigaword} which are generated based on global word-word co-occurrence statistics \cite{pennington2014glove}. For Twitter dataset, we use Twitter models  trained on 2 billion tweets.\footnote{http://nlp.stanford.edu/projects/glove/} We, then, create clusters of these word vectors using the K-means algorithm (K= 100, empirically chosen) for both datasets. Then, for each writing, we calculate the frequency of each cluster by checking the existence of each word of the document in the cluster. With this cluster based representation, we aim to capture the effect of semantically related words on the classification.
\subsubsection{Pragmatic Features}
Level of Committed Belief (LCB) \cite{Diab:2009:CBA:1698381.1698393,Prabhakaran:2010:ACB:1944566.1944683,werner-EtAl:2015:ExProM} encodes the linguistic modality of belief. It captures the belief commitment of a writer's or speaker's  to  propositions expressed in  text. All propositions are classified within a four-way belief type distinction: \emph{Committed Belief (CB)}: the writer believes the proposition; \emph{Non-committed Belief (NCB)}: the writer could believe the proposition but does not have a strong belief in it; \emph{Reported Belief (ROB)}: the writer reports on someone else's stated belief, whether or not they believe it; \emph{Non Attributable Belief (NA)}: the writer is not (or could not be) expressing a belief in the proposition. 
We hypothesize that schizophrenia patients may show more commitment than controls. Some patients with schizophrenia have delusions (fixed, false beliefs), which represent an excessive investment in an idea that in unaffected individuals would be brushed off. For example, coincidences can be over-interpreted in someone who is delusional, and delusion-proneness is known to be associated with cognitive biases and \emph{jumping to conclusions}. This may represent a greater level of committed belief, or tendency to believe things more strongly than those with a lower propensity toward delusions. Accordingly, we calculate the frequency of committed belief tags as features to the classifier. The differences between long-form text and Twitter data were significant enough that the LCB taggers did not perform sufficiently well on the Twitter data to allow meaningful analyses.

We also wanted to see whether patients exhibit more negative sentiment than controls. For that purpose, we use the Stanford Sentiment Analysis tool
\cite{Socher_recursivedeep}. Given a sentence, it predicts its sentiment at five possible levels: \emph{very negative, negative, neutral, positive,} and \emph{very positive}. For each writing, we  calculate the frequency of sentiment levels. Additionally, sentiment intensities are produced at the phrase level. Rather than categorical values, this intensity encodes the magnitude of the sentiment more explicitly. As such, we calculate the total intensity for each document as sum of its phrases' intensities at each level. For Twitter dataset, we use a sentiment classifier that was trained for social media data \cite{Columbia-Sentiment}. Its output includes three levels of sentiment \emph{negative, neutral, } and \emph{positive} without intensity information. 

\subsubsection{Feature Analysis}
To be able to better evaluate best performing features, we analyze them based on two feature selection algorithms: Information Gain (IG) for Random Forest and Recursive Feature Elimination (RFE) algorithm for SVM \cite{Guyon:2002:GSC:599613.599671}. The Information Gain measure selects the attributes that decrease the entropy the most. The RFE algorithm, on the other hand, selects features based on their weights based on the fact that the larger weights correspond to the more informative features.
\section{Results}
\label{sect:results}
\begin{figure*}[tp]
    \centering
           \subfloat[POS
        \label{fig:writingPOS}]{
        \includegraphics[scale=0.4]{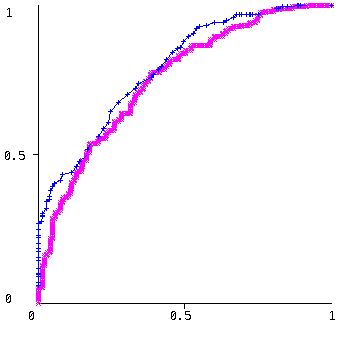}
    }
       \subfloat[Parse
        \label{fig:writingParse}]{
        \includegraphics[scale=0.4]{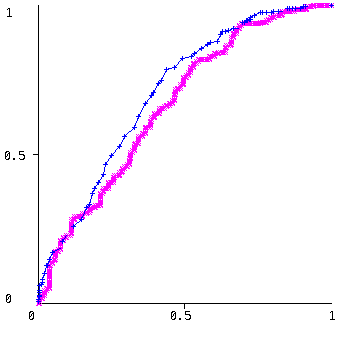} 
    }
     
     \subfloat[SRL
        \label{fig:writingSRL}]{
        \includegraphics[scale=0.4]{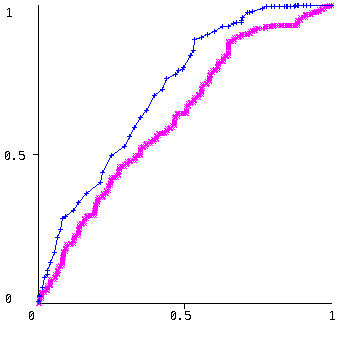}
    }
    \subfloat[Topics
        \label{fig:writingTopics}]{
        \includegraphics[scale=0.4]{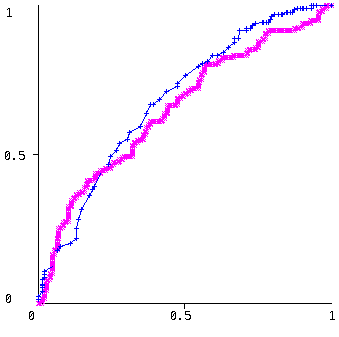} 
    }
     \subfloat[Clusters
        \label{fig:writingClusters}]{
        \includegraphics[scale=0.4]{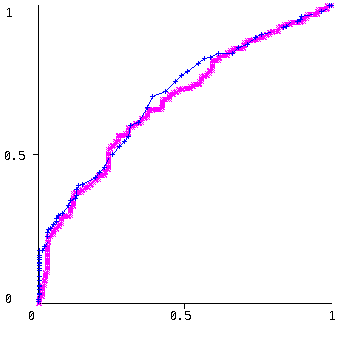} 
    }
    
        \subfloat[LCB
        \label{fig:writingCB}]{
        \includegraphics[scale=0.4]{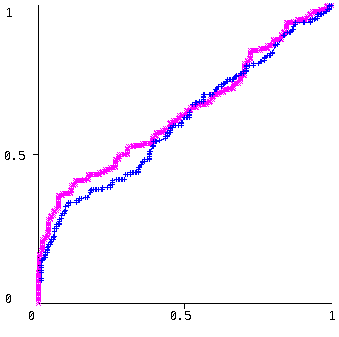}
    } 
    \subfloat[Sentiment
        \label{fig:writingSentiment}]{
        \includegraphics[scale=0.4]{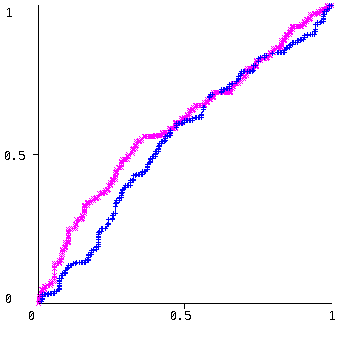} 
    } 
     \subfloat[Sentiment Intensity
        \label{fig:writingIntensity}]{
        \includegraphics[scale=0.4]{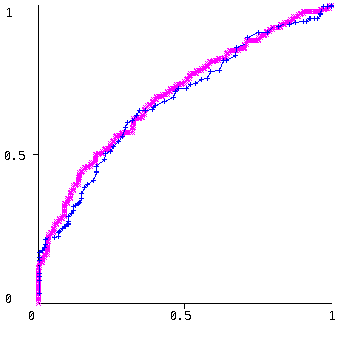} 
        }

    \caption{ROC Plots for LabWriting Dataset}
    \label{fig:writingROC}
\end{figure*}

\begin{figure*}[tp]
    \centering
   
    \subfloat[POS
        \label{fig:twitterPOS}]{
        \includegraphics[scale=0.45]{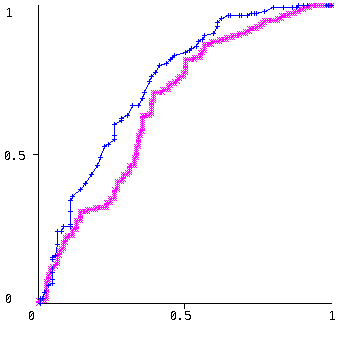}
    }
     \subfloat[Parse
        \label{fig:twitterParse}]{
        \includegraphics[scale=0.45]{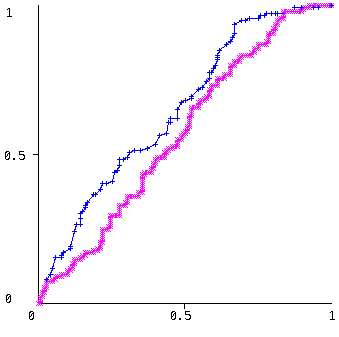} 
        }
      
         \subfloat[Topics
        \label{fig:twitterTopics}]{
        \includegraphics[scale=0.45]{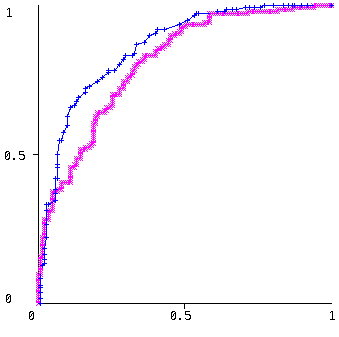} 
    }
       \subfloat[Clusters
        \label{fig:twitterClusters}]{
        \includegraphics[scale=0.45]{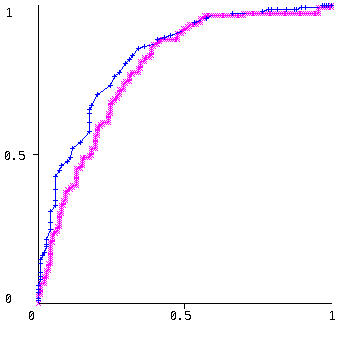}
    }
   
       \subfloat[Sentiment
        \label{fig:twitterSentiment}]{
        \includegraphics[scale=0.45]{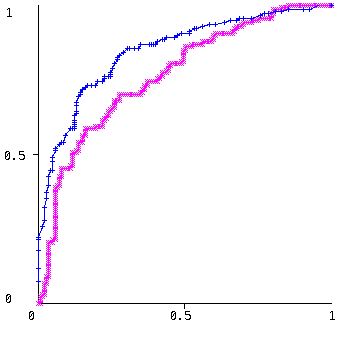} 
    }
    \caption{ROC Plots for Twitter Dataset}
    \label{fig:twitterROC}
\end{figure*}

The list of syntactic, semantic and pragmatic features are presented in Table \ref{tab:featureCat} for both datasets. Tables \ref{tab:classification} and \ref{tab:classificationTwitter} illustrate our results for the LabWriting dataset and Twitter dataset, respectively. The majority baseline F-Score is 34.39 for the LabWriting and 32.11 for Twitter dataset. The top performance for each dataset and classifier is shown in bold. The corresponding ROC plots for features are shown in Figures \ref{fig:writingROC} and \ref{fig:twitterROC} for LabWriting and Twitter datasets respectively. In each ROC plot, true positive rate (recall) is plotted against true negative rate where SVM is shown in magenta and Random Forest is shown in blue. The diagonal line from bottom left to upper right represents random guess and better performing results are closer to upper left corner. Overall, Random Forest performs better than SVM even though for some feature combinations, SVM's performance is higher. This could be due to bootstrapping of samples that takes place in Random Forest since both of the datasets are on the  smaller side.
\begin{table*}[thb]
\caption{\bf Feature Categories}
\label{tab:featureCat}
\begin{center}
    \begin{tabular}{| l | l | l | l |}
    \hline
    \bf{Category} & \bf{Writing Samples} & \bf{Twitter}  \\ \hline
    Syntactic  & POS, Dependency Parse & POS, Dependency Parse\\ \hline
     Semantic & SRL, Topics, Clusters & Topics , Clusters\\ \hline
         Pragmatic & LCB, Sentiment, Sentiment Intensity& Sentiment\\ \hline %
    \end{tabular}	
    \end{center}
\end{table*}

\begin{table*}[thb]
\caption{\bf Classification Performance of LabWriting Dataset}
\label{tab:classification}
{\centering \begin{tabular}
{lcccccccccl}
\hline
& \multicolumn{2}{c}{\bf{SVM}}& \multicolumn{2}{c}{\bf{Random Forest}} \\
\hline
\bf{Features} & \bf{AUC}&\bf{F-Score} & \bf{AUC} & \bf{F-Score}  \\
\hline
POS	&\bf{75.72}&68.48	&\bf{78.92}&69.76\\
Parse	&65.34&59.69&	66.68&65.94\\
SRL&	64.25&58.44&	70.62&64.93\\
Topics	&66.49&63.34&	68.26&62.55\\
Clusters	&69.68&64.98&	68.43&65.00\\
LCB	& 63.88&	60.52&63.07 &59.06\\
Sentiment	& 60.23 &54.40	&56.27&57.81 \\
Sentiment Intensity	&69.98&64.13&	69.39&64.96\\
Syntax	&74.17&\bf{68.88}&	75.78&68.67\\
Semantics	&66.46&61.32	&69.16&64.05 \\
Pragmatics	&68.95&62.56&	69.59&67.17 \\
Syntax + Semantics	&68.24&66.24&	76.60&\bf{70.29}\\
Syntax + Pragmatics	&73.13&66.36&	75.45&67.14  \\
Semantics + Pragmatics	&67.66&63.77	&71.84&65.22 \\ 
All&	70.23&66.29&	72.20&65.75\\ 
\hline
\end{tabular}
 \footnotesize \par}
\end{table*}
\begin{table*}[thb]
\caption{\bf Classification Performance of Twitter Dataset}
\label{tab:classificationTwitter}
{\centering \begin{tabular}
{lcccccccccl}
\hline
& \multicolumn{2}{c}{\bf{SVM}}& \multicolumn{2}{c}{\bf{Random Forest}} \\
\hline
\bf{Features} & \bf{AUC}& \bf{F-Score} & \bf{AUC}  &\bf{F-Score}  \\
\hline
POS	&69.34&63.19	&75.17&69.20\\
Parse&	58.63	&48.97&	63.72&	60.78\\
Topics	& 79.88&74.25	&83.48&79.07\\
Clusters	&78.02&72.55	&82.54&74.28\\
Sentiment	&75.79	&62.74	&85.28&79.75\\
Syntax	&74.87&65.01&	74.35&67.59\\
Semantics &	80.29&	70.57&	85.62&	74.86\\
Syntax+Semantics	&81.47&72.78	&86.35&78.16\\
Syntax+Pragmatics	&82.16	&73.88&83.55&75.70\\
Semantics+Pragmatics	&80.99&74.28	&\bf{88.98}&\bf{81.65}\\
All	&\bf{82.58}&\bf{74.78}&	88.01&	78.60\\
\hline
\end{tabular}
 \footnotesize \par}
\end{table*}


For LabWriting dataset, the best performing features according to F-Score are syntactic: POS+Parse (syntax) for SVM and syntax + semantics features for Random Forest. According to AUC, best performing feature is POS for both classifiers. For Twitter dataset, the best performing features according to both F-Score and AUC are the ones that include most of the combination of features: semantics + pragmatics for Random Forest and all features for SVM. 
Typically, essays, such as the ones in LabWriting dataset, are expected to have better syntax than informal tweets and as such syntactic features were not as predictive for tweets. We also analyze top performing features according to Information Gain measure and SVM RFE algorithm in Sections \ref{sect:topSyntax}, \ref{sect:topPragmatics}, \ref{sect:topSemantics} and explain the differences of results for the two datasets in Section \ref{sect:effectDataset}.
\subsection{Top Syntactic Features}
\label{sect:topSyntax}
Syntactic features perform well mainly for LabWriting dataset. Between POS tags and dependence parses, the former perform better for both datasets. For LabWriting dataset, the top POS tag is \emph{FW}, (Foreign Word). When we look at the words that were tagged \emph{FW}, they correspond to misspelled words. Even though this could be considered a criterion for schizophrenia patients, it may also depend on patients and controls' education and language skills which we expect it to be similar but it may still show some differences. 
  Another top  POS tag is \emph{LS}, (List item marker), which was assigned to small case \emph{i} which in reality refers to pronoun \emph{I}. This could imply that the patients prefer to talk about themselves. 
This coincides with several other studies \cite{doi:10.1080/02699930441000030,chung2007social} which found that use of first person singular is associated with negative affective states such as depression. 
Because of the likelihood of comorbidity of mental illnesses, this requires further investigation as to whether this is specific to schizophrenia patients or not. Finally, another top POS tag is \emph{RP}, adverbial particle and top parse tag is \emph{advmod}, adverb modifier. This could mean the ratio of adverbs used could be a characteristic of patients. 
Finally for Twitter dataset, the top POS tag is \emph{\#} corresponding to hash tags. This could be an important discriminative feature between patients and controls as patients use less hashtags than controls.
\subsection{Top Semantic Features}
\label{sect:topSemantics}
For classification using semantic features, clusters, topics and SRL perform comparably.  For LabWriting dataset, top SRL features consist of general categories and some specific ones that could be relevant for schizophrenia patients. General labels are \emph{Quantity} and \emph{Social Event}. 
More specific labels are \emph{Morality Evaluation, Catastrophe, Manipulate into Doing} and \emph{Being Obligated}. Words that are labeled as such are listed in Table \ref{tab:srlFeatures}. These two different sets of  labels could be due to the type of questions asked to the patients. One question is neutral in nature talking about their daily life whereas the other is about the things that make them angry and more emotionally charged. A second semantic feature is the topic distributions of writings. The top words from the most informative topics are listed in Table \ref{tab:topicFeatures}. For LabWriting dataset, one of the top topics consist of words about typical Sunday activities corresponding to one of the questions asked. The second top topic, on the other hand, consist of words that show the anger of the author. For Twitter dataset, one of the topics consist of schizophrenia-related words and the other consist of hate words. Again, the top topics seem to contain relevant information in analyzing schizophrenia patients' writings and classification using topic features perform comparably well. As a final semantic feature, we use dense cluster features. The classification performance of cluster features is similar to classification performance using topics. However, cluster features' analysis is not as interpretable as topics, since they are formed from massive online data resources.
\begin{table*}[thb]
\small
\caption{\bf Top Discriminative SRL Features}
\label{tab:srlFeatures}
\begin{center}
    \begin{tabular}{| l | l | l | }
    \hline
    \bf{Label} & \bf{Sample Words/Phrases} \\ \hline
    Quantity & several, both, all, a lot, many, a little, a few, lots \\ \hline
    Social Event & social, dinner, picnics, hosting, dance \\ \hline
    Morality Evaluation  &wrong, depraved, foul, evil, moral\\ \hline
    Catastrophe & incident,tragedy, suffer
\\ \hline
     Manipulate into Doing &harassing, bullying\\ \hline
     Being Obligated & duty, job, have to, had to, should, must, responsibility, entangled,  task, assignment
\\ \hline
    \end{tabular}
    \end{center}
\end{table*}
\begin{table*}[thb]
\small
\caption{\bf Discriminative Topics' Top Words}
\label{tab:topicFeatures}
\begin{center}
    \begin{tabular}{| l | l | l | l | }
    \hline
    \bf{Method} & \bf{Dataset} & \bf{Top Words} \\ \hline
    IG  &Writing& church sunday wake  god service pray sing worship bible spending attend  thanking \\ \hline
    IG \& RFE &  Writing& i'm can't trust upset person lie real feel honest lied lies judge lying guy steal
\\
\hline
    IG \& RFE  &Twitter&   god  jesus mental schizophrenic schizophrenia illness paranoid truth medical  evil  \\ \hline 
    IG \& RFE  &Twitter& don love people fuck life feel fucking hate shit stop god person sleep bad girl die 
\\ \hline
    \end{tabular}
    \end{center}
\end{table*}
\subsection{Top Pragmatic Features}
\label{sect:topPragmatics}
When it comes to pragmatic features, top sentiment features are \emph{neutral}, \emph{negative} and \emph{very negative} (LabWriting only). For sentiment intensity, \emph{neutral intensity}, \emph{negative intensity} and \emph{very negative intensity} are more informative which is consistent with sentiment categorical analysis. In general, neutral sentiment is the most common for a given text and for patients, we would expect to see more negative sentiment and this was confirmed by this analysis. However, negative sentiment could also be prominent in other psychiatric diseases such as post-traumatic stress disorder (PTSD)\cite{coppersmith15a}, as such, by itself, it may not be a discriminatory feature for schizophrenia patients. For classification purposes, sentiment intensity features performed better than sentiment features. This could be due to the fact that intensity values are more specific and collected at word/phrase level in contrast to sentence level. 
For LabWriting, committed belief analysis returned the \emph{CB} tag as most informative. This confirms with our hypothesis that schizophrenia patients may show more commitment of their belief to the propositions expressed in their writings. However, for classification purposes, LCB features were not as successful. This may be due to a genre  mismatch between patients' writing in either LabWriting/Twitter and the LCB tagger trained models. We plan to adapt the LCB tagger to handle less formal writings for future analysis.
\subsection{Effect of Datasets' Characteristics}
\label{sect:effectDataset}
The two datasets have some commonalities and differences and present different challenges. The LabWriting dataset was collected in a more controlled manner and follows a structure that can be expected from a short essay. Accordingly, NLP tools applied to  these writings are successful. On the other hand, the Twitter dataset consists of combinations of short text that include many abbreviations that are not standard, e.g. users' own solutions to fixed length limit imposed by Twitter. It is also very informal in nature and thus lacks proper grammar and syntax more frequently than LabWriting. Hence, some machine learning approaches for NLP analysis of these tweets are limited even though social media specific tools were used such as POS tagger \cite{Gimpel:2011:PTT:2002736.2002747}, dependency parser \cite{kong2014dependency}, sentiment analysis tool \cite{Columbia-Sentiment}, and Twitter models for dense clusters. For instance, even though we were able to compute POS tags and parse trees for tweets, the tag set is much smaller than PennTree Bank tags set.

Similarly, some approaches such as LCB and SRL were not successful on tweets. On the other hand, both datasets consist of patients and controls with similar demographics (age, gender, etc), thus we largely expect patients and controls to have similar linguistic capabilities. In addition, for LabWriting dataset, patients and controls were recruited from the same neighborhoods. We have no such explicit guarantees for the Twitter dataset, though they were excluded if they did not primarily tweet in English. Accordingly, any differentiation these classification methods found can largely be attributed to the illness.  Finally, LabWriting dataset had many spelling errors. We elected not to employ any spelling correction techniques (since misspelling may very well be a feature meaningful to schizophrenia). This likely negatively influenced the calculation of some of the features which depend on correct spelling such as SRL, and LCB analysis. 
\section{Related Work}
\label{sec:background}
To date, some studies have investigated applying Latent Semantic Analysis (LSA) to the problem \cite{Elvevag:2007aa} of lexical coherence and they found significant distinctions between schizophrenia patients and controls. The work of \cite{Bedi:2015aa} extends this approach by incorporating syntax, i.e., phrase level LSA measures and POS tags. In the latter related work, several measures based on LSA representation were developed to capture the possible incoherence in patients. In our study, we used LDA to capture possible differences in themes between patients and controls. LDA is a more descriptive technique than LSA since topics are represented as distributions over vocabulary and top words for topics provide a way to understand the theme that they represent. We also incorporated syntax to our analysis with POS tags and additionally dependency parses. Another work by \cite{Howes:2013aa} predicts outcomes by analyzing doctor-patient communication in therapy using LDA. Even though manual analysis of LDA topics with manual topics seem promising, classification using topics does not perform as successful unless otherwise additional features are incorporated such as doctors' and patients' information. Although, we had detailed demographic information for LabWriting dataset and derived age and sex information for Twitter dataset, we chose not to incorporate them to the classification process be able focus solely on writing's characteristics. 

The work of Mitchell et al. \cite{mitchell-hollingshead-coppersmith:2015:CLPsych} is, in many respects, similar to ours by examining schizophrenia using LDA, clustering and sentiment analysis. Their sentiment analysis is lexicon-based using Linguistic Inquiry Word Count (LIWC) \cite{Tausczik_thepsychological} categories. In our approach to sentiment analysis, we utilized a machine learning approach. Lexicon-based approaches generally have higher precision at the cost of lower recall. Having coverage of more of the content may be beneficial for analysis and interpretation, so we opt to use a more generalizable machine learning approach. For clustering, they used Brown clustering; whereas, we used clusters trained on global word vectors which were learned from large amounts of online data. This has the advantage that we could capture words and/or semantics that may not be learned from our dataset. Finally, their use of LDA is similar to our approach, i.e. representing documents as topic distributions, and their analysis does not include syntactic and dense cluster features. They had their best performance with an accuracy value of 82.3 using a combination of topic based representation and their version of sentiment features. In our analysis, combination of semantic and pragmatic features performed the best with an accuracy value of 81.7. Due to possible differences in preprocessing, parameter selection, and randomness that exist in the experiments, the results are not directly comparable, however, this also shows that the difficulty of applying more advanced machine learning based NLP techniques for Twitter dataset.

Finally, the work of Hong et al. \cite{HongKMPN12,hong2015lexical} analyzes small set of narratives based on five emotions. In this work, a very large amount of features are considered including LIWC: all lexical with the exception of POS language model. They apply various feature selection mechanisms including RFE to get to fewer most predictive features. In our study, we preferred a machine learning approach to sentiment for better recall and generalization instead of manually created dictionaries such as LIWC. Also, our features are more diverse to capture various linguistic characteristics: syntax, semantics and pragmatics. 
 
\section{Conclusion}
\label{sect:Conclusion}
Computational assessment models of schizophrenia may provide ways  for clinicians to monitor symptoms more effectively and 
a deeper understanding of schizophrenia and the underpinning cognitive biases could benefit affected individuals, families, and society at large.
Objective and passive assessment of schizophrenia symptoms (e.g., delusion or paranoia) may provide clarity to clinical assessments, which currently rely on patients' self-reporting symptoms. Furthermore, the techniques discussed here hold some potential for early detection of schizophrenia.
This would be greatly beneficial to young people and first-degree relatives of schizophrenia patients who are prodromal (clinically appearing to be at high risk for schizophrenia) but not yet delusional/psychotic, since it would allow targeted early interventions.

Among the linguistic features considered for this study, syntactic features provide the biggest boost in classification performance for LabWriting dataset. For Twitter dataset, combination of features such as semantics and pragmatics for SVM and syntax, semantics and pragmatics for Random Forest have the best performance. 

In the future, we will be focusing on the features that showed the most promise in this study and also add new features such as committed belief for pragmatics. Finally, we are collecting more data and we will expand our analysis to more mental health datasets.
\bibliography{acl2017-efsun}
\bibliographystyle{elsarticle-num}
\end{document}